%% file: paper.tex
\documentclass{article}

\usepackage{arxiv}

\usepackage[utf8]{inputenc} 
\usepackage[T1]{fontenc}    
\usepackage{hyperref}       
\usepackage{url}            
\usepackage{booktabs}       
\usepackage{amsmath,amsfonts}       
\usepackage{nicefrac}       
\usepackage{microtype}      
\usepackage{lipsum}
\usepackage{xcolor}
\usepackage{listings}
\usepackage{bm}
\usepackage{graphicx}
\usepackage[linesnumbered,vlined]{algorithm2e}

\definecolor{codegreen}{rgb}{0,0.5,0}
\definecolor{codepurple}{rgb}{0.88,0,0.32}
\definecolor{backcolour}{rgb}{0.97,0.97,0.97}
 
\lstdefinestyle{coding}{
    backgroundcolor=\color{backcolour},   
    commentstyle=\color{codegreen},
    keywordstyle=\color{codepurple},
    basicstyle=\ttfamily,
    breakatwhitespace=false,         
    breaklines=true,                 
    captionpos=b,                    
    keepspaces=true,                 
    numbers=none,                    
    numbersep=5pt,                  
    showspaces=false,                
    showstringspaces=false,
    showtabs=false,                  
    tabsize=4
}
 
\title{Learnergy: Energy-based Machine Learners}

\author{
  Mateus Roder, Gustavo Henrique de Rosa, João Paulo Papa \\
  Department of Computing \\
  São Paulo State University \\
  Bauru, São Paulo - Brazil \\
  \texttt{\{mateus.roder, gustavo.rosa, joao.papa\}@unesp.br} \\
}
\begin{document}

\maketitle

\begin{abstract}
Throughout the last years, machine learning techniques have been broadly encouraged in the context of deep learning architectures. An exciting algorithm denoted as Restricted Boltzmann Machine relies on energy- and probabilistic-based nature to tackle the most diverse applications, such as classification, reconstruction, and generation of images and signals. Nevertheless, one can see they are not adequately renowned compared to other well-known deep learning techniques, e.g., Convolutional Neural Networks. Such behavior promotes the lack of researches and implementations around the literature, coping with the challenge of sufficiently comprehending these energy-based systems. Therefore, in this paper, we propose a Python-inspired framework in the context of energy-based architectures, denoted as Learnergy. Essentially, Learnergy is built upon PyTorch to provide a more friendly environment and a faster prototyping workspace and possibly the usage of CUDA computations, speeding up their computational time.
\end{abstract}


\input{sections/introduction.tex}
\input{sections/literature.tex}
\input{sections/theory.tex}
\input{sections/learnergy.tex}
\input{sections/library.tex}
\input{sections/applications.tex}
\input{sections/conclusion.tex}

\section*{Acknowledgments}
The authors are grateful to S\~ao Paulo Research Foundation (FAPESP) grants \#2013/07375-0, \#2014/12236-1, \#2019/02205-5 and \#2019/07825-1, as well as CNPq grants \#307066/2017-7 and \#427968/2018-6.

\bibliographystyle{unsrt}  
\bibliography{paper}

\end{document}

%% file: sections/introduction.tex
\section{Introduction}
\label{s.intro}

Artificial Intelligence (AI) has achieved considerable attention in the last years, mainly due to its capacity to explore and solve previously unsolvable problems. Pattern Recognition, Computer Vision, Natural Language Processing are some areas that have benefited from the advancements in AI and were able to achieve state-of-the-art results in a wide variety of tasks.

Another exciting topic denoted as Machine Learning deals explicitly with the design of algorithms that solve AI problems. The advancement of machine learning research enabled the resolution of common-daily issues, such as image classification, object recognition, text summarization, autonomous cars, biometric identification, among others. Furthermore, the compulsion of creating human alike models and solving computer-vision related tasks enhanced traditional machine learning methods into more sophisticated techniques, known as deep learning~\cite{BengioTPAMI:13}.

Nevertheless, machine learning and deep learning are highly susceptible to the data that it is feed with, being only reasonable when the dataset is accurately modeled, representing the real possibilities of the problem~\cite{Schmidt:18}. An additional problem is when the dataset is not balanced, providing more samples pertain to one class than another(s). Such a problem may vitiate the learning process and prompt the technique to learn the most common class instead of adequately learning the whole dataset.

A recent technique that has attracted big spotlights is the Restricted Boltzmann Machine~\cite{Hinton:02} (RBM), mainly due to its simple architecture and vast learning capability. The Restricted Boltzmann Machine is a stochastic network that deals with probabilities and physical concepts, such as entropy and energy. Moreover, it can learn real data distributions and reconstruct the original data in different latent spaces~\cite{Srivastava:12}, i.e., acting as an auto encoder-decoder and extracting new features from the data distribution. Notwithstanding, energy-based systems usually suffer from overfitting under the lack of data, causing premature convergence and poor generalization over unseen data. Different approaches can be highlighted to overcome such a problem, such as regularization, data augmentation, and hyperparameter fine-tuning.

One can perceive that there is a lack of energy-based open-sourced frameworks in the literature. There are only a few implementations that deal with RBMs, but none of them in a framework format. Moreover, they are inconsistent among themselves, implementing distinct learning algorithms from the one provided by Hinton et al.~\cite{Hinton:02}. Therefore, in this paper, we propose an open-source energy-based Machine Learning framework, called Learnergy\footnote{https://github.com/gugarosa/learnergy}. Essentially, the idea is to provide a user-friendly environment to work with energy-based architectures by creating high-level methods and classes, removing from the user the burden of programming at a mathematical level.  The main contributions of this paper are threefold: (i) to introduce an energy-based machine learning library in the Python language, (ii) to provide an easy-to-go implementation and user-friendly framework, and (iii) to fill the lack of research regarding energy-based machine learning techniques.

The remnant of this work is presented as follows. Section~\ref{s.literature} discusses a revision over the literature and some relevant frameworks in the context of energy-based machine learning techniques. Section~\ref{s.theory} presents a theoretical background concerning Restricted Boltzmann Machines, Dropout-based Restricted Boltzmann Machines, and Deep Belief Networks. Section~\ref{s.learnergy} introduces an overview of the Learnergy library, such as its architecture and included packages. Section~\ref{s.library} introduces more thorough concepts regarding Learnergy usage, such as installation, documentation, included examples, and unitary tests. Additionally, Section~\ref{s.applications} presents an outline on how to work with the library, i.e., how to run predefined examples, and how to model a new learning procedure. Finally, Section~\ref{s.conclusion} describes conclusions and future works.

%% file: sections/literature.tex
\section{Literature Review and Related Frameworks}
\label{s.literature}

Restricted Boltzmann Machines are fruitful approaches that attempt to tackle supervised and unsupervised problems. They offer an energy-based architecture inspired by probabilistic concepts capable of learning the probabilistic distribution of the training data and further interpreting unseen data. They are employed in a wide range of applications, such as collaborative filtering~\cite{Salakhutdinov:07}, image reconstruction~\cite{Nair:10}, classification~\cite{Larochelle:08}, denoising~\cite{Tang:12}, data generation~\cite{Ranzato:10}, among others. For instance, Srivastava et al.~\cite{Srivastava:12} found out that RBMs are capable of learning real data distributions and reconstructing the original data in different latent spaces, i.e., acting as an auto encoder-decoder and extracting new features from the data distribution. Yosinski and Lipson~\cite{Yosinski:12} highlighted some approaches for visualizing the behavior of a Restricted Boltzmann Machine during its learning procedure and provided an overview concerning its complexities comprehension. Thornton et al. proposed to address the RBM complexity using auto-learning tools, which combine parameter fine-tuning with feature selection techniques~\cite{Thornton:13}. Moreover, Li et al.~\cite{Li:16} proposed an improved version of the RBM, the so-called ``temperature-based RBM" (T-RBM), which employs a new temperature parameter during the learning process to manipulate the neurons' activation.

Although some works in the literature foster the Restricted Boltzmann Machines, one might find it challenging to search for open-sourced frameworks. There are only a few implementations, such as \url{https://github.com/yell/boltzmann-machines}, \url{https://github.com/echen/restricted-boltzmann-machines}, \url{https://github.com/artem-oppermann/Restricted-Boltzmann-Machine}, and \url{https://github.com/GabrielBianconi/pytorch-rbm}, but none of them seems to be in a framework format, only loosely code that implements the RBMs. Additionally, there is some inconsistency between these packages, especially throughout the learning algorithm, where some of them use distinct versions from the one provided by Hinton et al.~\cite{Hinton:02}. Moreover, they lack documentation and test cases, which help users understand the code and implement new methods and classes. Apart from that, we could find a C-based framework, denoted as LibDEEP\footnote{https://github.com/jppbsi/LibDEEP}. Nevertheless, as the library is implemented in C language, it is not very easy to integrate with other frameworks or packages and use CUDA-based operations.

Therefore, Learnergy attempts to fill the gaps concerning energy-based machine learning frameworks. It is completely implemented in Python and PyTorch, providing optimized performance, CUDA-capable operations, and several gimmicks that help the computational burden. Furthermore, it has comprehensive documentation, test cases, several pre-loaded examples, commented lines, continuous integration, full-time maintenance, and support.

\clearpage

%% file: sections/theory.tex
\section{Theoretical Background}
\label{s.theory}

Before plunging into Learnergy's library, we present a theoretical background about the Restricted Boltzmann Machines, Dropout Restricted Boltzmann Machines, and Deep Belief Networks.

\subsection{Restricted Boltzmann Machines}
\label{ss.rbm}

Restricted Boltzmann Machines are stochastic neural networks based on energy principles guided by physical laws and characterized by energy, entropy, and temperature factors. Most of the time, these networks learn in an unsupervised fashion and apply to a wide variety of problems that range from image reconstruction, collaborative filtering, and feature extraction to pre-training deeper networks.

\begin{figure}[!h]
\centering
\includegraphics[scale=0.75]{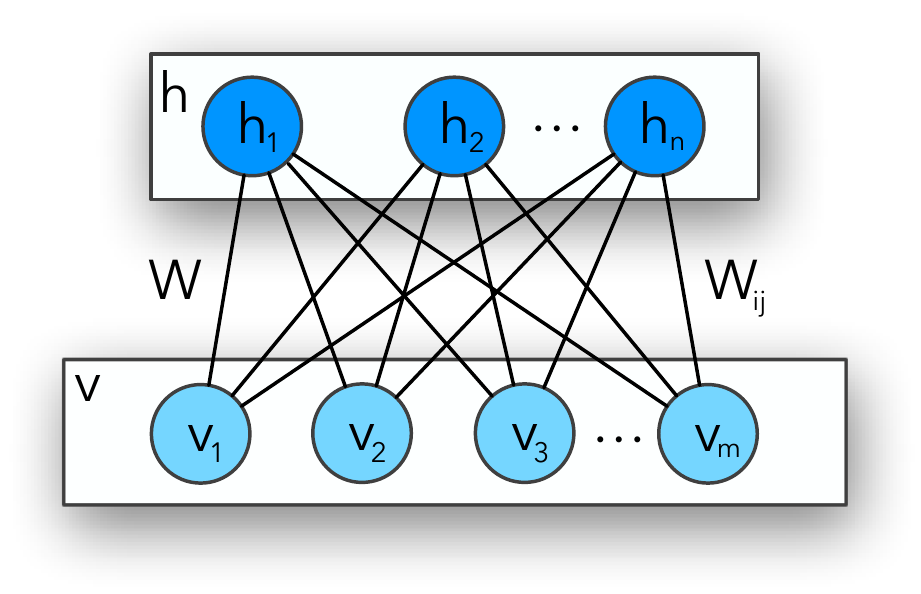}
\caption{Typical RBM network.}
\label{f.rbm}
\end{figure}

In terms of architecture, RBM contain a visible layer $\mathbf{v}$ with $m$ units and a hidden layer $\mathbf{h}$ with $n$ units. Additionally, a real-valued matrix $\mathbf{W}_{m\times n}$ models the weights between the visible and hidden neurons, where $w_{ij}$ represents the connection between the visible unit $v_i$ and the hidden unit $h_j$. Figure~\ref{f.rbm} describes the well-know vanilla RBM architecture.

Let $\mathbf{v}\in\{0,1\}^m$ and $\mathbf{h}\in\{0,1\}^n$ be the binary visible and hidden units, respectively. The RBM energy function can be formulated as follows:

\begin{equation}
\label{e.energy_rbm}
E(\mathbf{v},\mathbf{h})=-\sum_{i=1}^ma_iv_i-\sum_{j=1}^nb_jh_j-\sum_{i=1}^m\sum_{j=1}^nv_ih_jW_{ij},
\end{equation}
where $\mathbf{a}$ and $\mathbf{b}$ are the biases of visible and hidden units, respectively. Furthermore, one can compute the probability of a configuration $(\mathbf{v},\mathbf{h})$ as follows:

\begin{equation}
\label{e.probability_configuration}
P(\mathbf{v},\mathbf{h})=\frac{e^{-E(\mathbf{v},\mathbf{h})}}{\displaystyle\sum_{\mathbf{v},\mathbf{h}}e^{-E(\mathbf{v},\mathbf{h})}},
\end{equation}
where the denominator normalizes the equation, standing for all possible configurations involving the visible and hidden units. Also, the probabilities of these units are updated as follows:

\begin{equation}
\label{e.probh}
P(h_j=1|\mathbf{v})=\sigma\left(\sum_{i=1}^mW_{ij}v_i+b_j\right),
\end{equation}
and, 
\begin{equation}
\label{e.probv}
P(v_i=1|\mathbf{h})=\sigma\left(\sum_{j=1}^nW_{ij}h_j+a_i\right),
\end{equation}
where $\sigma$ stands for the well-known sigmoid function. Essentially, an RBM learning algorithm pursues in estimating the values of $\mathbf{W}$, $\mathbf{a}$ and $\mathbf{b}$.

\subsection{Dropout Restricted Boltzmann Machines}
\label{ss.rbm_dropout}

\begin{figure}[!h]
\centering
\includegraphics[scale=0.75]{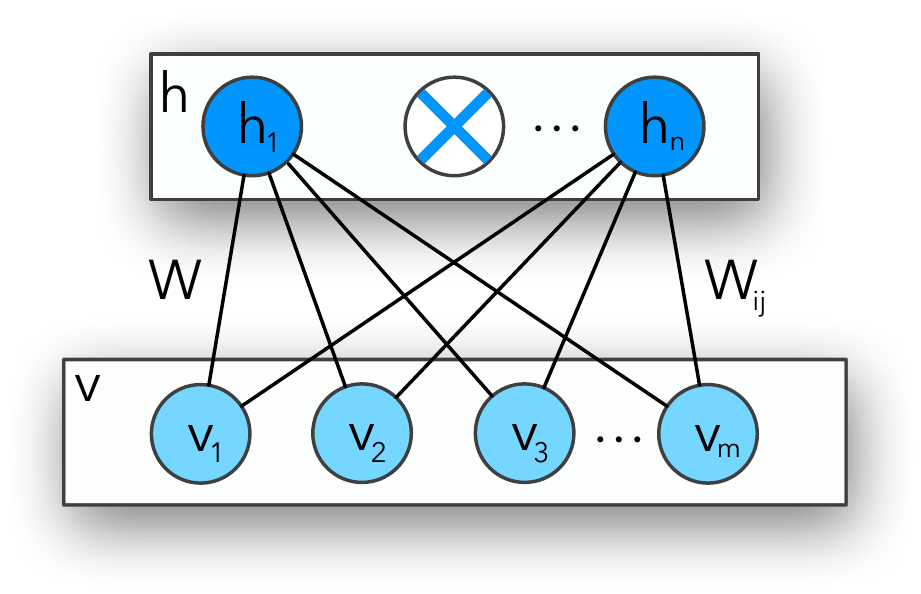}
\caption{The Dropout-based RBM architecture.}
\label{f.rbm_dropout}
\end{figure}

Considering the RBM model described in Section~\ref{ss.rbm}, it is possible to extend it in a Dropout RBM with a simple binary random vector $\mathbf{r} \in \{0, 1\}^n$. In this new architecture, $\mathbf{r}$ stands for the activation of neurons in the hidden layer, where each variable $r_i$ contains the value $0$ (zero) with probability $p$, independent of other variables $r_j$, where $i\neq j$. If $r_i$ equals to $0$ (zero), the hidden unit $h_i$ is temporarily dropped along with its connections, otherwise it is held. Figure~\ref{f.rbm_dropout} illustrates this procedure, where unit $h_2$ is shutoff.

Notice that $\mathbf{r}$ is sampled from a Bernoulli distribution~\cite{Srivastava:14}, which is re-sampled for every training batch. Therefore, a Dropout RBM network can be seen as a blend of several RBMs, each using distinct subsets of their hidden layers.

As units were dropped from the hidden layer, Equation~\ref{e.probh} can be rewritten as:

\begin{equation}
\label{e.probh_dropout}
P(h_j=1|\mathbf{r}, \mathbf{v})=
   \begin{cases}
      0, & \text{if}\ r_j=0 \\
      \sigma\left(\sum_{i=1}^mW_{ij}v_i+b_j\right), & \text{otherwise}.
   \end{cases}
\end{equation}

As we are training our model with different subsets, our weight matrix $\mathbf{W}$ needs to be scaled at testing time, being multiplied by $p$ in order to adjust its weights. In short, after learning its parameters, the new weight matrix $\mathbf{W}'$ is obtained as:

\begin{equation}
    \mathbf{W}' = p\mathbf{W}.
\end{equation}

\subsection{Deep Belief Networks}
\label{ss.dbn}

\begin{figure}[!ht]
\centering
\includegraphics[scale=0.75]{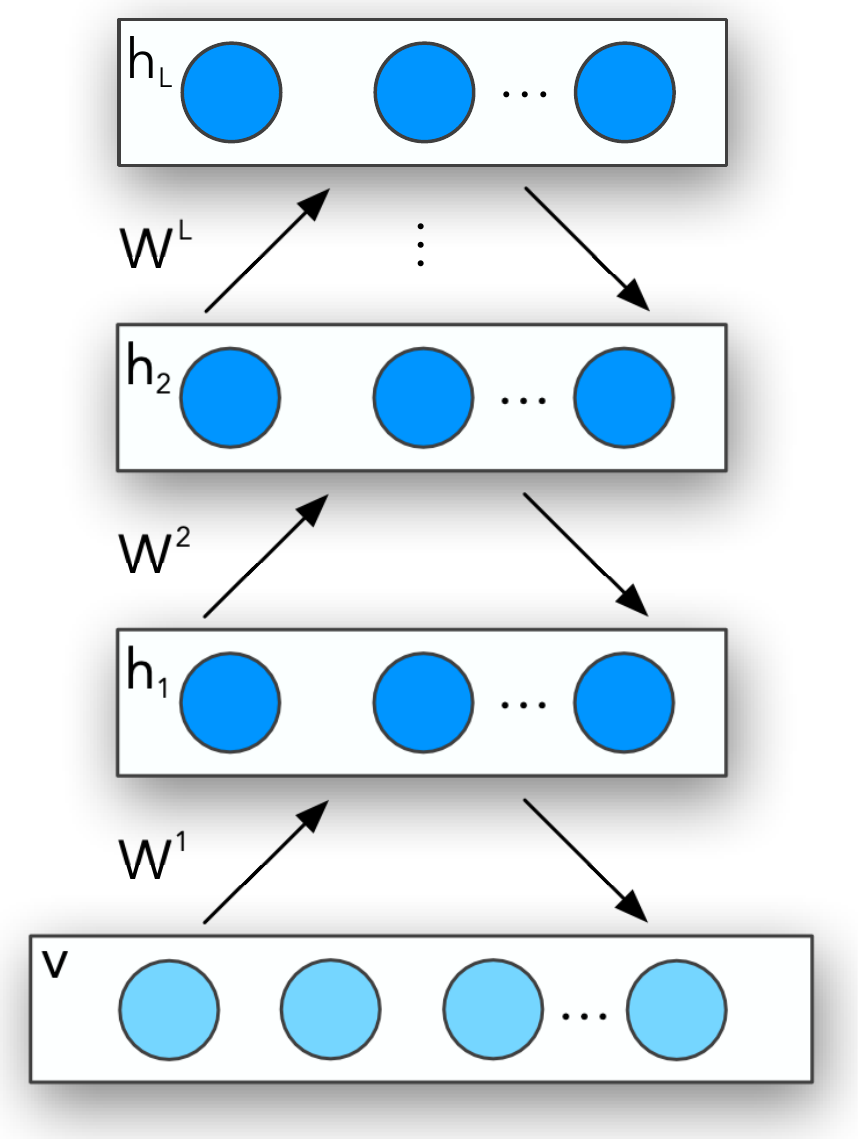}
\caption{The standard DBN architecture.}
\label{f.dbn}
\end{figure}

Restricted Boltzmann Machines can be used as building blocks to create the so-called Deep Belief Networks~\cite{Hinton:06}. In short, they are composed of a group of stacked RBMs and trained greedily, i.e., each RBM does not acknowledge others throughout its learning process. Figure~\ref{f.dbn} describes a DBN architecture, where each RBM at a specific layer is the one portrayed in Figure~\ref{f.rbm}.

One can observe that the DBN is a model composed of a visible and $L$ hidden layers, being $\mathbf{W}^i$ the weight matrix of an RBM at layer $i$. Additionally, the hidden units at layer $i$ are converted to input units in the layer $i+1$. Although Figure~\ref{f.dbn} does not illustrate, there are also bias units for the visible and hidden layers.

The learning method introduced by Hinton et al.~\cite{Hinton:06} uses a fine-tuning procedure after training each RBM independently. Such optimization can be accomplished through Gradient Descent or Backpropagation algorithms. The optimization algorithm minimizes a fitness function, usually an error measure based on an extra layer's output on top of the DBN's architecture. Such a layer can be formed by logistic or softmax units in a classification task.

\clearpage

%% file: sections/learnergy.tex
\section{Learnergy}
\label{s.learnergy}

\begin{figure}[!ht]
\centering
\includegraphics[scale=0.4]{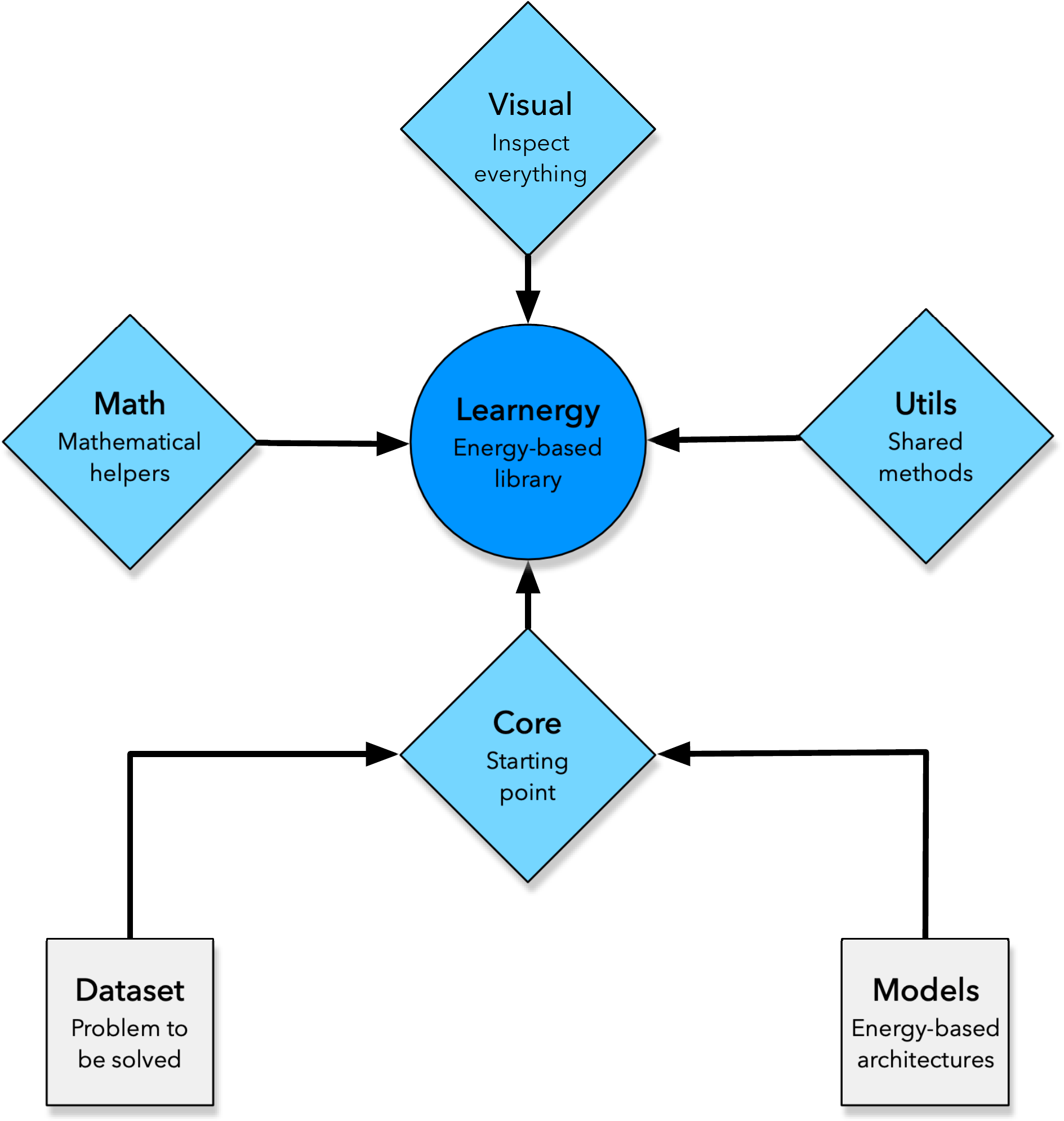}
\caption{Flowchart of Learnergy's architecture.}
\label{f.flowchart}    
\end{figure}

Learnergy's structure is split into several packages, where each implements specific methods and classes. Figure~\ref{f.flowchart} represents a summary of Learnergy's architecture, while the next sections present each of its packages within more details.

\subsection{Core}
\label{ss.core}

\begin{figure}[!ht]
\centering
\includegraphics[scale=0.5]{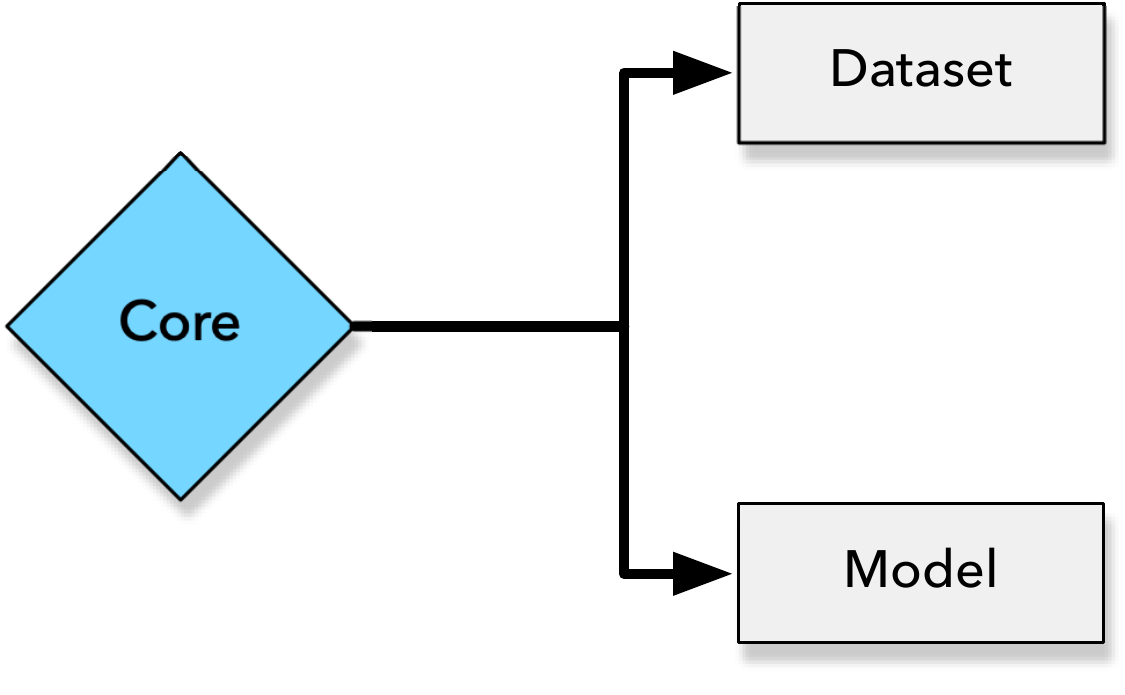}
\caption{Flowchart of Learnergy's core package.}
\label{f.flowchart_core}    
\end{figure}

The core package implements all of Learnergy's parent classes. In other words, it serves as a building block for other classes, which may be needed when creating more complex structures. As portrayed in Figure~\ref{f.flowchart_core}, two modules compose the core package, as follows:

\begin{itemize}

\item \textbf{Dataset:} The dataset is an extension of PyTorch's class to deal with every possible input to the energy-based architectures;

\item \textbf{Model:} The model is an extension of PyTorch's class with additional features, such as saving an object containing the training's history.

\end{itemize}

\subsection{Math}
\label{ss.math}

\begin{figure}[!ht]
\centering
\includegraphics[scale=0.5]{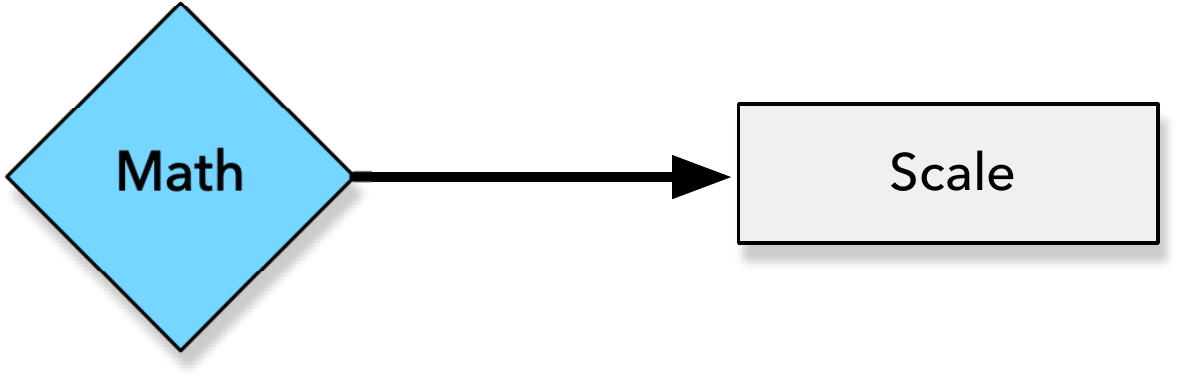}
\caption{Flowchart of Learnergy's math package.}
\label{f.flowchart_math}    
\end{figure}

Learnergy also offers a mathematical package in an attempt to facilitate the user's prototype. It contains low-level implementations, such as the ones illustrated by Figure~\ref{f.flowchart_math}. Typically, some repeated functions that re-used throughout the library can be found in this package, such as:

\begin{itemize}

\item \textbf{Scale:} Common-use scaling functions are defined in this module.

\end{itemize}

\subsection{Models}
\label{ss.models}

\begin{figure}[!ht]
\centering
\includegraphics[scale=0.5]{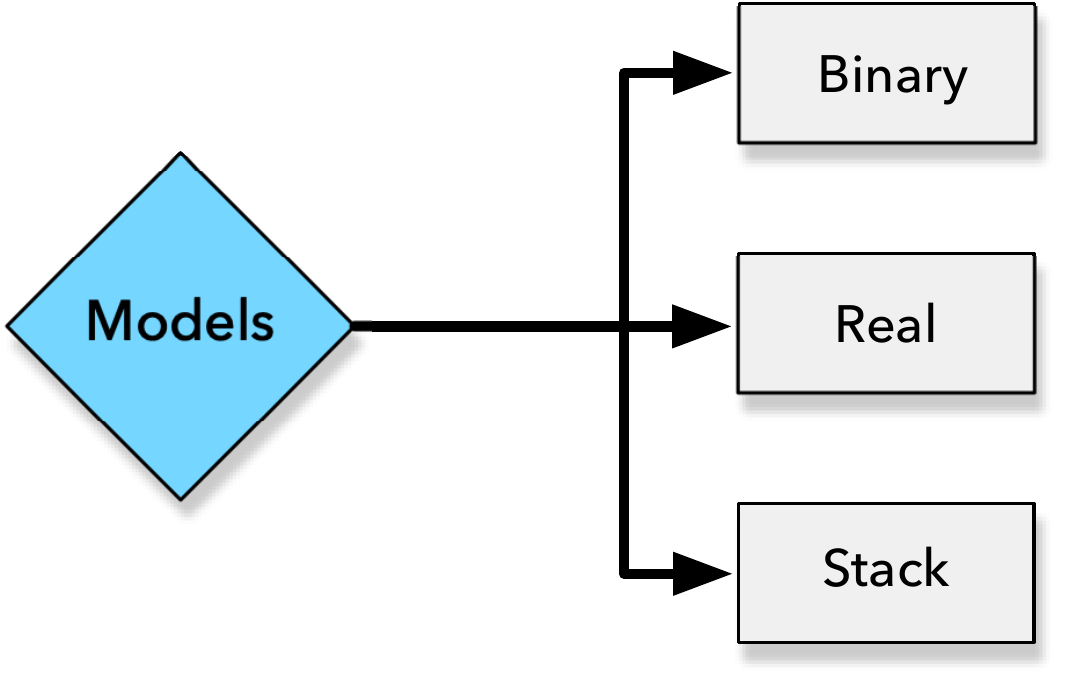}
\caption{Flowchart of Learnergy's models package.}
\label{f.flowchart_models}    
\end{figure}

There are several approaches to be conducted when designing an energy-based architecture. Therefore, the models' package provides the necessary blocks that compose these high-level abstractions. Currently, Learnergy offers three sub-packages of energy-based architectures, which are illustrated by Figure~\ref{f.flowchart_models}.

\subsubsection{Binary}
\label{sss.binary}

\begin{itemize}

\item \textbf{ConvRBM:} Convolutional Restricted Boltzmann Machine~\cite{Lee:09};

\item \textbf{DiscriminativeRBM:} Discriminative Bernoulli-Bernoulli Restricted Boltzmann Machines~\cite{Larochelle:08};

\item \textbf{DropoutRBM:} Dropout Restricted Boltzmann Machines~\cite{Srivastava:14};

\item \textbf{EDropoutRBM:} Energy-based Dropout Restricted Boltzmann Machines;

\item \textbf{HybridDiscriminativeRBM:} Hybrid Discriminative Bernoulli-Bernoulli Restricted Boltzmann Machines~\cite{Larochelle:08};

\item \textbf{RBM:} Bernoulli-Bernoulli Restricted Boltzmann Machines~\cite{Hinton:02}.

\end{itemize}

\subsubsection{Real}
\label{sss.real}

\begin{itemize}

\item \textbf{GaussianConvRBM:} Gaussian-based Convolutional Restricted Boltzmann Machines~\cite{Lee:09};

\item \textbf{GaussianRBM:} Gaussian-Bernoulli Restricted Boltzmann Machines~\cite{Hinton:02};

\item \textbf{GaussianReluRBM:} Gaussian ReLU Restricted Boltzmann Machines~\cite{Hinton:12};

\item \textbf{SigmoidRBM:} Sigmoid-Bernoulli Restricted Boltzmann Machines~\cite{Hinton:12};

\item \textbf{VarianceGaussianRBM:} Gaussian-Bernoulli Restricted Boltzmann Machines with Adaptive Variance~\cite{Hinton:02}.
	
\end{itemize}

\subsubsection{Stack}
\label{sss.stack}

\begin{itemize}

\item \textbf{DBN:} Deep Belief Networks~\cite{Hinton:06};

\item \textbf{ResidualDBN:} Residual-based Deep Belief Networks.

\end{itemize}

\subsection{Utils}
\label{ss.utils}

\begin{figure}[!ht]
\centering
\includegraphics[scale=0.5]{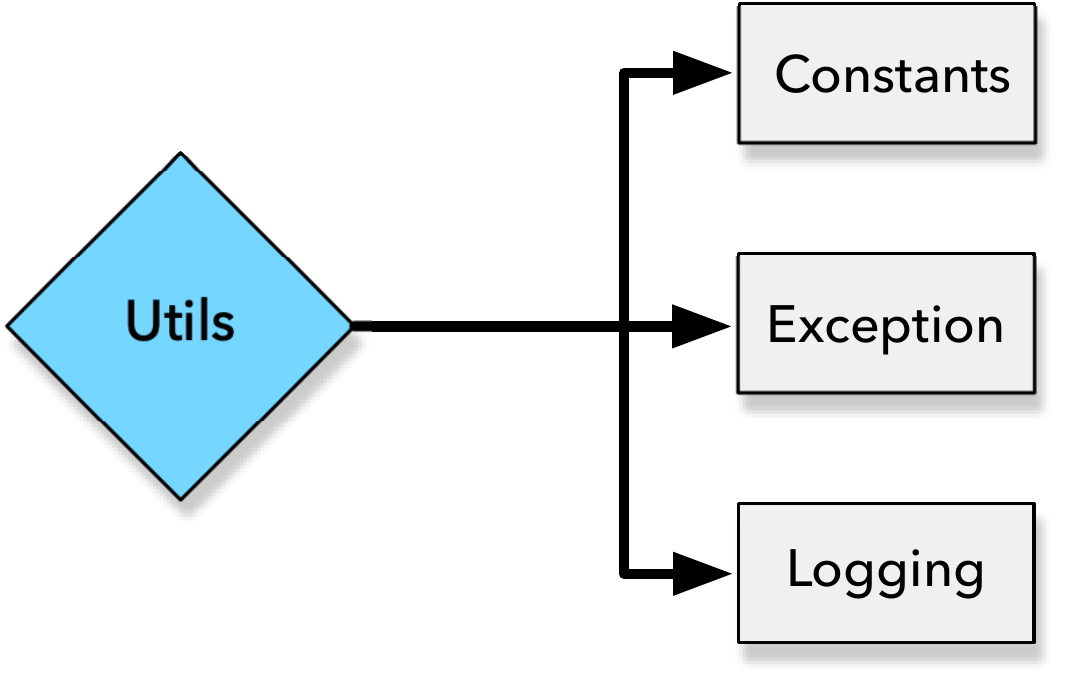}
\caption{Flowchart of Learnergy's utils package.}
\label{f.flowchart_utils}    
\end{figure}

This package offers common functions that are shared across the library, as shown in Figure~\ref{f.flowchart_utils}. The following modules are implemented in this package:

\begin{itemize}

\item \textbf{Constants:} Constants are fixed values that do not alter across the library;

\item \textbf{Exception:} An exception module implements common errors and exceptions that might happen when invalid arguments are used;

\item \textbf{Logging:} Every invoked method is saved into a log file. The log helps detect potential errors, warnings, or even success messages.

\end{itemize}

\subsection{Visual}
\label{ss.visual}

\begin{figure}[!ht]
\centering
\includegraphics[scale=0.5]{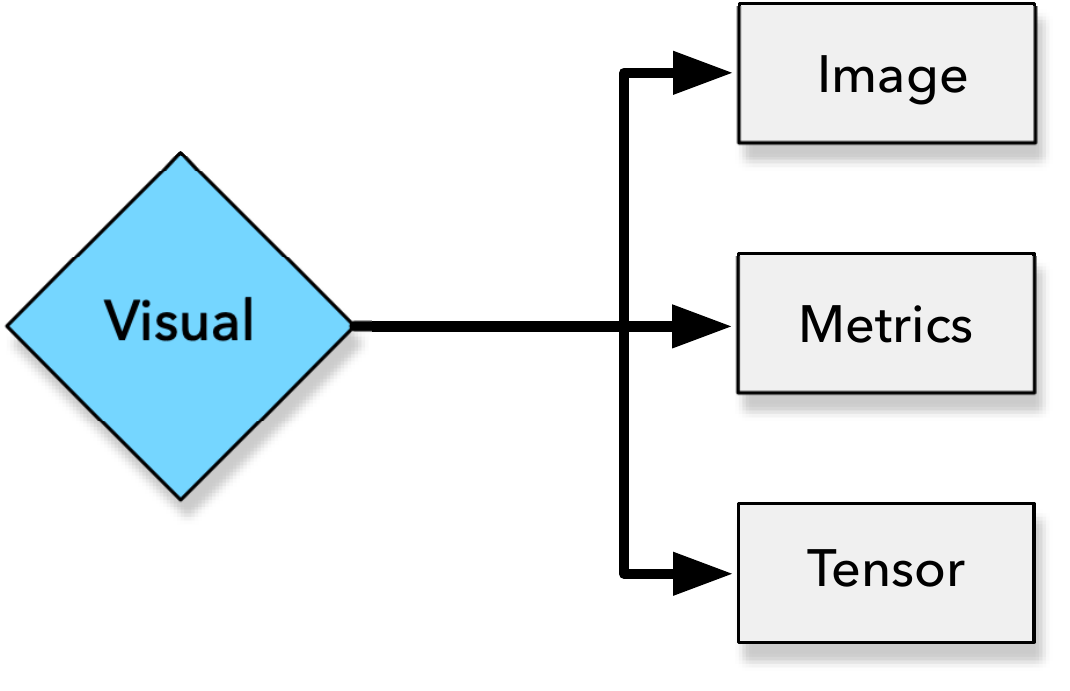}
\caption{Flowchart of Learnergy's visual package.}
\label{f.flowchart_visual}    
\end{figure}

Finally, we provide a visual-related package to plot mosaics, reconstructed images, or furnish convergence graphics. The visual package implements the following modules, which are also portrayed in Figure~\ref{f.flowchart_visual}:

\begin{itemize}

\item \textbf{Image:} Provides image-based methods for creating mosaics from the model's weights;

\item \textbf{Metrics:} Allows users to input their history object in order to furnish metrics convergence graphics;

\item \textbf{Tensor:} Inspects a tensor and re-creates an image of it.

\end{itemize}

%% file: sections/library.tex
\section{Using the Library}
\label{s.library}

This section explains how to install the Learnergy library and the first actions to start working with it. Mainly, one can read its documentation or make usage of the already-included examples. Furthermore, there are implemented methods that handle unitary tests and assess if everything is running as expected.

\subsection{Installation}
\label{ss.installation}

First of all, we understand that things should be smooth instead of complicated. Therefore, Learnergy will eternally be the one-to-go package, from the very first installation to its future usage. Just perform the following command under the most favored Python environment (standard, conda, virtualenv):

\verb|pip install learnergy|

As a matter of choice, it is possible to use the bleeding-edge version by cloning its repository and installing it:

\verb|git clone https://github.com/gugarosa/learnergy.git|

\verb|pip install .|

Learnergy's single dependency is the PyTorch package, making it possible to be installed everywhere, despite the machine's operational system.

\subsection{Documentation}
\label{ss.documentation}

We present a fully documented reference\footnote{https://learnergy.readthedocs.io}, including everything the library offers, to fulfill the desire for additional comprehension. One can understand within more depth the thoughts and policies behind Learnergy, such as its elementary classes, methods, and more. Therefore, Learnergy's documentation is the definitive source to study how the library was produced or even learn how to improve it with contributions.

\subsection{Available Examples}
\label{ss.example}

In the \verb|examples/| folder, we present example scripts for every package the library implements, as follows:

\begin{itemize}

\item \textbf{Core:} \verb|create_dataset.py|, \verb|create_model.py|;
\item \textbf{Math:} \verb|unitary_scaling.py|;
\item \textbf{Models:} \verb|binary/create_conv_rbm.py|, \verb|binary/create_discriminative_rbm.py|, \\ \verb|binary/create_dropout_rbm.py|, \verb|binary/create_e_dropout_rbm.py|, \\ \verb|binary/create_hybrid_discriminative_rbm.py|, \verb|binary/create_rbm.py|, \\ \verb|real/create_gaussian_conv_rbm.py|, \verb|real/create_gaussian_rbm.py|, \\ \verb|real/create_gaussian_relu_rbm.py|, \verb|real/create_sigmoid_rbm.py|, \\ \verb|real/create_variance_gaussian_rbm.py|, \verb|stack/create_dbn.py|, \verb|stack/create_residual_dbn.py|;

\item \textbf{Visual:} \verb|create_weights_mosaic.py|, \verb|plot_metrics.py|, \verb|show_reconstructed_sample.py|.

\end{itemize}

Each example presents high-level descriptions of predefined classes and methods. In other words, there are conventional explanations regarding how to instantiate each class and choose the most proper arguments.

\subsection{Test Engines}
\label{ss.test}

\begin{figure}[!ht]
\centering
\includegraphics[scale=0.775]{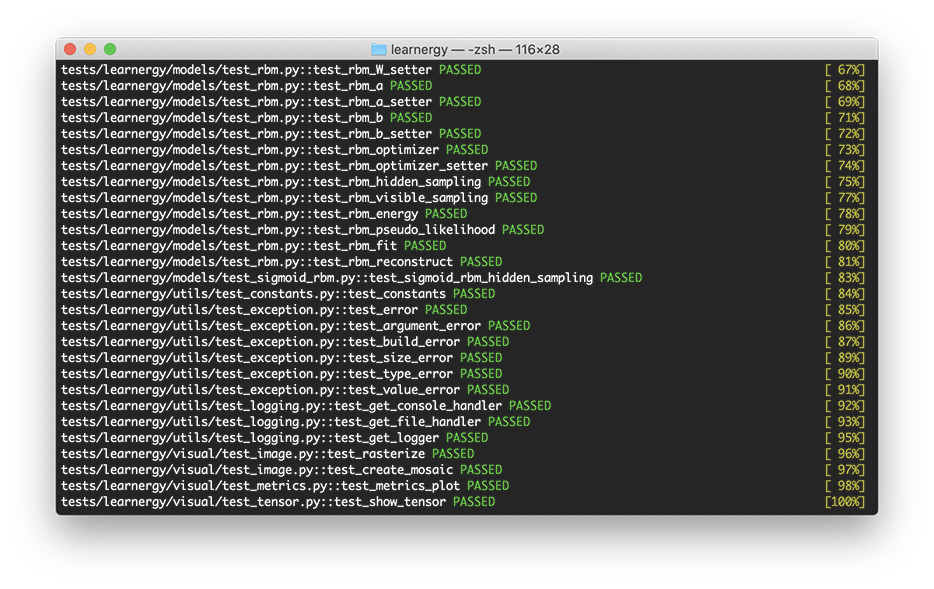}
\caption{Example of outputs generated by a PyTest running.}
\label{f.run_tests}    
\end{figure}

\begin{figure}[!ht]
\centering
\includegraphics[scale=0.775]{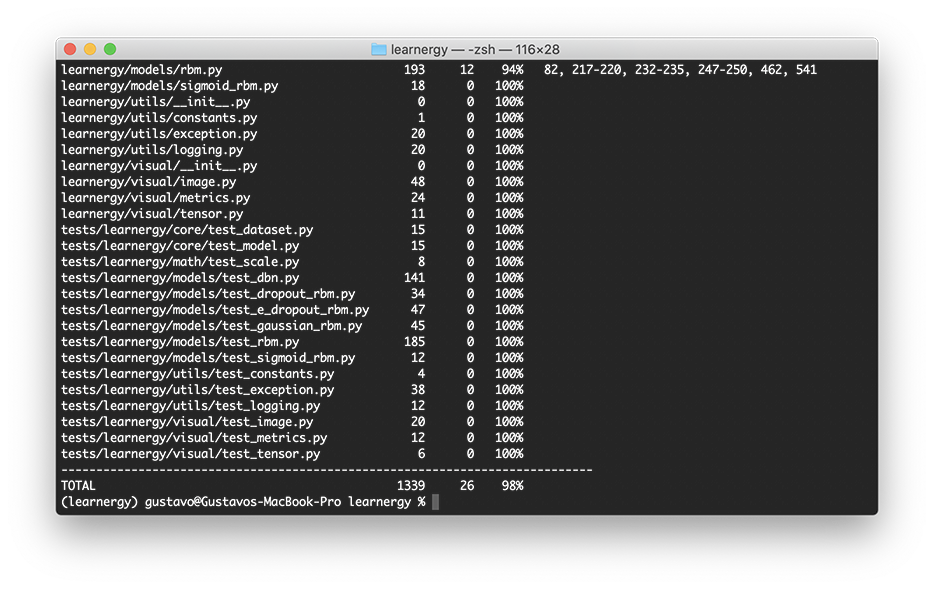}
\caption{Example of code coverage by a Coverage running.}
\label{f.coverage_tests}    
\end{figure}

Tests are designed to provide a robust analysis of the code. Therefore, Learnergy implements a variety of tests to verify whether everything is running as required or not. Currently, we offer two ways to conduct the tests:

\begin{itemize}
    \item \textbf{PyTest:} By running the command \verb|pytest tests/|, it performs the implemented tests and returns a log showing whether they succeeded or failed, as portrayed by Figure~\ref{f.run_tests}; 

    \item \textbf{Coverage:}, An additional extension to PyTest is the coverage module. Despite being almost identical to PyTest, it also outputs a report stating the code coverage, as represented by Figure~\ref{f.coverage_tests}. Its utilization is simple: \verb|coverage run -m pytest tests/| and \verb|coverage report -m|. 
\end{itemize}

%% file: sections/applications.tex
\section{Practical Applications}
\label{s.applications}

This section demonstrates how to perform an energy-based learning task with Learnergy and quickly define the eight pre-loaded applications incorporated within the library.

\subsection{Initial Steps}
\label{ss.initial_steps}

It is straightforward to use Learnergy's packages after installation. One can find in the \verb|examples/applications| folder eight rudimentary examples which show key points in the library implementations, as follows:

\begin{itemize}

\item \textbf{Convolutional RBM classification:} \verb|binary/conv_rbm_classification.py|;
\item \textbf{Convolutional RBM training:} \verb|binary/conv_rbm_training.py|;
\item \textbf{Discriminative RBM training:} \verb|binary/discriminative_rbm_training.py|;
\item \textbf{Dropout RBM training:} \verb|binary/dropout_rbm_training.py|;
\item \textbf{Energy-based Dropout RBM training:} \verb|binary/e_dropout_rbm_training.py|;
\item \textbf{Hybrid Discriminative RBM training:} \verb|binary/hybrid_discriminative_rbm_training.py|;
\item \textbf{RBM classification:} \verb|binary/rbm_classification.py|;
\item \textbf{RBM training:} \verb|binary/rbm_training.py|;
\item \textbf{Gaussian Convolutional RBM classification:} \verb|real/gaussian_conv_rbm_classification.py|;
\item \textbf{Gaussian Convolutional RBM training:} \verb|real/gaussian_conv_rbm_training.py|;
\item \textbf{Gaussian RBM training:} \verb|real/gaussian_rbm_training.py|;
\item \textbf{Gaussian ReLU RBM training:} \verb|real/gaussian_relu_rbm_training.py|;
\item \textbf{Sigmoid RBM training:} \verb|real/sigmoid_rbm_training.py|;
\item \textbf{Variance Gaussian RBM training:} \verb|real/variance_gaussian_rbm_training.py|;
\item \textbf{DBN classification:} \verb|stack/dbn_classification.py|;
\item \textbf{DBN training:} \verb|stack/dbn_training.py|;
\item \textbf{Residual DBN training:} \verb|stack/residual_dbn_training.py|;
\item \textbf{Loading a pre-trained architecture:} \verb|loading_pre_trained_model.py|.

\end{itemize}

\begin{figure}[!ht]
\centering
\includegraphics[scale=0.75]{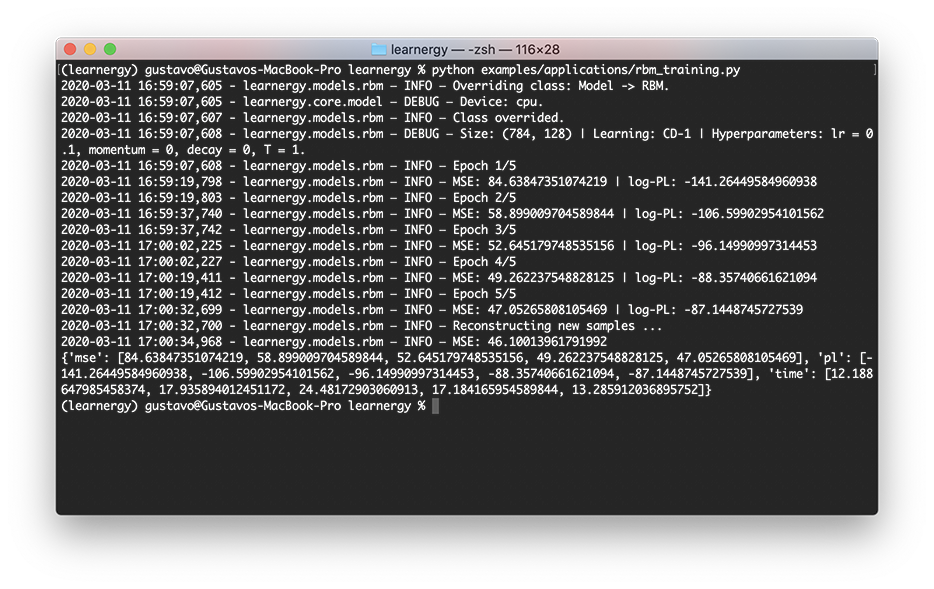}
\caption{Output logs generated by executing a Learnergy learning procedure.}
\label{f.get_started}    
\end{figure}

Each example comprises the following pipeline: loading the dataset, instantiating an architecture, fitting the training data, and reconstructing the testing data. Finally, after performing the learning process, it is possible to save the model in a disk-file for further inspection. Figure~\ref{f.get_started} illustrates the output information originated by a Learnergy execution.

The distinction between the provided scripts consists of the type of architecture. As for now, we offer seven distinct energy-based architectures, e.g., DBN, Dropout RBM, Energy-based Dropout RBM, Gaussian RBM, RBM, Sigmoid RBM, and Variance Gaussian RBM. Additionally, we offer a script for loading pre-trained models and performing further analysis.

\clearpage

\subsection{Modeling a Learning Procedure}
\label{ss.function}

Some standard rules should be followed in order to model a new learning procedure. Firstly, the data should be loaded, which in this case, we will be loading a common dataset known as MNIST:

\begin{lstlisting}[language=Python]
import torchvision

# Creating training and testing dataset
train = torchvision.datasets.MNIST(root='./data', train=True, download=True, transform=torchvision.transforms.ToTensor())
test = torchvision.datasets.MNIST(root='./data', train=False, download=True, transform=torchvision.transforms.ToTensor())    
\end{lstlisting}

Afterward, we can instantiate an RBM architecture:

\begin{lstlisting}[language=Python]
from learnergy.models.binary import RBM

# Creating an RBM
model = RBM(n_visible=784, n_hidden=128, steps=1, learning_rate=0.1, momentum=0, decay=0, temperature=1, use_gpu=True)    
\end{lstlisting}

Finally, we can fit the architecture and perform new reconstructions:

\begin{lstlisting}[language=Python]
# Training an RBM
mse, pl = model.fit(train, batch_size=128, epochs=5)

# Reconstructing test set
rec_mse, v = model.reconstruct(test, batch_size=10000)    
\end{lstlisting}

One can also persist in the model to disk using PyTorch's saving function or even check its learning history.

\begin{lstlisting}[language=Python]
import torch

# Saving model
torch.save(model, 'model.pth')

# Checking the model's history
print(model.history)    
\end{lstlisting}

%% file: sections/conclusion.tex
\section{Conclusions}
\label{s.conclusion}

This article presents an open-source Python-inspired library for handling energy-based machine learning architectures, known as Learnergy. Based upon an object-oriented criterion, Learnergy implements a modern yet straightforward framework, allowing users to prototype new energy-based architectures speedily. The library implements a wide variety of energy-based architectures and additional classes and functions that assist the architectures' workflow. Moreover, Learnergy provides a model-saving method, which can be used to pre-train models and retrieve knowledgeable insights about learning performance. Concerning future works, we plan to offer more energy-based architectures, such as Deep Boltzmann Machines, and a more robust visual package, which will allow users to supply their saved models and furnish improved charts. Furthermore, we aim at improving our implementations by better exploring PyTorch's mechanisms and possible bottlenecks.

%% file: paper.bbl
\begin{thebibliography}{10}

\bibitem{BengioTPAMI:13}
Y.~Bengio, A.~Courville, and P.~Vincent.
\newblock Representation learning: A review and new perspectives.
\newblock {\em IEEE Transactions on Pattern Analysis and Machine Intelligence},
  35(8):1798--1828, 2013.

\bibitem{Schmidt:18}
L.~Schmidt, S.~Santurkar, D.~Tsipras, K.~Talwar, and A.~Madry.
\newblock Adversarially robust generalization requires more data.
\newblock In {\em Advances in Neural Information Processing Systems}, pages
  5014--5026, 2018.

\bibitem{Hinton:02}
G.~Hinton.
\newblock Training products of experts by minimizing contrastive divergence.
\newblock {\em Neural Computation}, 14(8):1771--1800, 2002.

\bibitem{Srivastava:12}
N.~Srivastava and R.~Salakhutdinov.
\newblock Multimodal learning with deep boltzmann machines.
\newblock In {\em Advances in neural information processing systems}, pages
  2222--2230, 2012.

\bibitem{Salakhutdinov:07}
R.~Salakhutdinov, A.~Mnih, and G.~Hinton.
\newblock Restricted boltzmann machines for collaborative filtering.
\newblock In {\em Proceedings of the 24th international conference on Machine
  learning}, pages 791--798. ACM, 2007.

\bibitem{Nair:10}
V.~Nair and G.~Hinton.
\newblock Rectified linear units improve restricted boltzmann machines.
\newblock In {\em Proceedings of the 27th international conference on machine
  learning (ICML-10)}, pages 807--814, 2010.

\bibitem{Larochelle:08}
H.~Larochelle and Y.~Bengio.
\newblock Classification using discriminative restricted boltzmann machines.
\newblock In {\em Proceedings of the 25th international conference on Machine
  learning}, pages 536--543, 2008.

\bibitem{Tang:12}
Y.~Tang, R.~Salakhutdinov, and G.~Hinton.
\newblock Robust boltzmann machines for recognition and denoising.
\newblock In {\em 2012 IEEE Conference on Computer Vision and Pattern
  Recognition}, pages 2264--2271. IEEE, 2012.

\bibitem{Ranzato:10}
M.~Ranzato, A.~Krizhevsky, and G.~Hinton.
\newblock Factored 3-way restricted boltzmann machines for modeling natural
  images.
\newblock In {\em Proceedings of the Thirteenth International Conference on
  Artificial Intelligence and Statistics}, pages 621--628, 2010.

\bibitem{Yosinski:12}
J.~Yosinski and H.~Lipson.
\newblock Visually debugging restricted boltzmann machine training with a 3d
  example.
\newblock In {\em Representation Learning Workshop, 29th International
  Conference on Machine Learning}, 2012.

\bibitem{Thornton:13}
C.~Thornton, F.~Hutter, H.~Hoos, and K.~Leyton-Brown.
\newblock Auto-weka: Combined selection and hyperparameter optimization of
  classification algorithms.
\newblock In {\em Proceedings of the 19th ACM SIGKDD international conference
  on Knowledge discovery and data mining}, pages 847--855. ACM, 2013.

\bibitem{Li:16}
G.~Li, L.~Deng, Y.~Xu, C.~Wen, W.~Wang, J.~Pei, and L.~Shi.
\newblock Temperature based restricted boltzmann machines.
\newblock {\em Scientific reports}, 6:19133, 2016.

\bibitem{Srivastava:14}
N.~Srivastava, G.~Hinton, A.~Krizhevsky, I.~Sutskever, and R.~Salakhutdinov.
\newblock Dropout: A simple way to prevent neural networks from overfitting.
\newblock {\em The Journal of Machine Learning Research}, 15(1):1929--1958,
  January 2014.

\bibitem{Hinton:06}
G.~Hinton, S.~Osindero, and Y.~Teh.
\newblock A fast learning algorithm for deep belief nets.
\newblock {\em Neural Computation}, 18(7):1527--1554, 2006.

\bibitem{Lee:09}
Honglak Lee, Roger Grosse, Rajesh Ranganath, and Andrew~Y Ng.
\newblock Convolutional deep belief networks for scalable unsupervised learning
  of hierarchical representations.
\newblock In {\em Proceedings of the 26th annual international conference on
  machine learning}, pages 609--616, 2009.

\bibitem{Hinton:12}
G.E. Hinton.
\newblock A practical guide to training restricted boltzmann machines.
\newblock In G.~Montavon, G.~Orr, and K.~M{\"u}ller, editors, {\em Neural
  Networks: Tricks of the Trade}, volume 7700 of {\em Lecture Notes in Computer
  Science}, pages 599--619. Springer Berlin Heidelberg, 2012.

\end{thebibliography}
